\documentclass[conference]{IEEEtran}

\usepackage{bbm}
\usepackage{hyperref}
\usepackage{graphicx}
\usepackage{graphicx}
\usepackage{caption}
\usepackage{subcaption}
\usepackage{booktabs}
\usepackage{algorithmic}
\usepackage{cite}
\usepackage{textcomp}
\usepackage{xcolor}
\usepackage{url}
\usepackage{amsmath,amssymb,amsfonts}
\def\BibTeX{{\rm B\kern-.05em{\sc i\kern-.025em b}\kern-.08em
    T\kern-.1667em\lower.7ex\hbox{E}\kern-.125emX}}

\newcommand{\etal}{{\em et~al.}}

\newcommand{\RNum}[1]{\uppercase\expandafter{\romannumeral #1\relax}}

\begin{document}

\title{MCD: A Model-Agnostic Counterfactual Search Method For Multi-modal Design Modifications}

\author{\IEEEauthorblockN{Lyle Regenwetter}
\IEEEauthorblockA{
\textit{Massachusetts Institute of Technology}\\
Cambridge, MA \\
regenwet@mit.edu}
\and
\IEEEauthorblockN{Yazan Abu Obaideh}
\IEEEauthorblockA{
\textit{ProgressSoft}\\
Amman, Jordan \\
yazan.amer@protonmail.com}
\and
\IEEEauthorblockN{Faez Ahmed}
\IEEEauthorblockA{
\textit{Massachusetts Institute of Technology}\\
Cambridge, MA \\
faez@mit.edu}
}
% \SetAuthors{%
% 	Lyle Regenwetter\affil{1}\CorrespondingAuthor{regenwet@mit.edu}, 
% 	Yazan Abu Obaideh\affil{2}, 
% 	Faez Ahmed\affil{1}
% 	}

% \SetAffiliation{1}{Massachusetts Institute of Technology, Cambridge, MA }
% \SetAffiliation{2}{ProgressSoft, }

\maketitle

\begin{abstract}

Designers may often ask themselves how to adjust their design concepts to achieve demanding functional goals. To answer such questions, designers must often consider counterfactuals, weighing design alternatives and their projected performance. This paper introduces Multi-objective Counterfactuals for Design (MCD), a computational tool that automates and streamlines the counterfactual search process and recommends targeted design modifications that meet designers' unique requirements. MCD improves upon existing counterfactual search methods by supporting multi-objective requirements, which are crucial in design problems, and by decoupling the counterfactual search and sampling processes, thus enhancing efficiency and facilitating objective trade-off visualization. The paper showcases MCD's capabilities in complex engineering tasks using three demonstrative bicycle design challenges. In the first, MCD effectively identifies design modifications that quantifiably enhance functional performance, strengthening the bike frame and saving weight. In the second, MCD modifies parametric bike models in a cross-modal fashion to resemble subjective text prompts or reference images. In a final multidisciplinary case study, MCD tackles all the quantitative and subjective design requirements introduced in the first two problems, while simultaneously customizing a bike design to an individual rider's biomechanical attributes. By exploring hypothetical design alterations and their impact on multiple design objectives, MCD recommends effective design modifications for practitioners seeking to make targeted enhancements to their designs\footnote{The code, test problems, and datasets used in the paper are available to the public at \hyperref[decode.mit.edu/projects/counterfactuals/]{decode.mit.edu/projects/counterfactuals/}}.

\end{abstract}

\section{Introduction}
Modifying existing designs to generate new ones is an essential aspect of various engineering sectors, such as aerospace, automotive, architecture, pharmaceuticals, consumer goods, and many others. 
Design modification significantly impacts the performance, efficiency, and safety of engineered systems. Effective methods for design modification can lead to more sustainable and environmentally friendly technologies, better transportation systems, and safer infrastructure. Furthermore, improved design modification methods can enable cost savings and improved efficiency, making products more accessible and affordable for society. However, coming up with good design modifications can be challenging, as it requires navigating huge design spaces and making numerous trade-offs between competing objectives. Often, there are too many design attributes and potential modifications to consider. Not surprisingly, designers often struggle with the available choices.

As a designer, the ability to ask ``What if?'' questions is crucial in the iterative process of design modification. By exploring hypothetical scenarios, designers can identify opportunities to improve design performance and functionality. However, answering ``What if?'' questions can be challenging as it requires considering an extensive range of potential modifications and their effects on multiple design objectives. Such questions can be addressed by a powerful reasoning tool called a counterfactual, which allows a designer to explore a hypothetical design modification and its impact on multiple design objectives. A counterfactual is a hypothetical situation that depicts what could have happened if a specific event or action did not occur. It requires envisioning an alternate reality where a different choice or decision was made and analyzing the differences in results. Counterfactuals are often employed in reasoning, decision-making, and causal inference. They aid in comprehending the impact of particular events or actions on outcomes and considering the ramifications of various choices. 

Counterfactuals are typically employed to understand how an outcome would change given a different set of actions. This style of counterfactual can be applied to design problems to answer questions like: ``How would the performance of this design change if I modified this particular attribute?'' There are many tools to predict these `forward' counterfactuals, such as simulations and predictive models. In this work, we instead consider `inverse' counterfactuals, which ask: ``What actions would be needed to result in this other outcome?'' In design contexts, this often equates to the question: ``What attributes of my design would I need to change to achieve a particular requirement, such as a performance target, design classification, or functionality?'' Such questions can be answered through `inverse counterfactual search.' 

This paper proposes a novel approach for inverse counterfactual search using an AI-driven multi-objective optimization. Our proposed approach, Multi-objective Counterfactuals for Design (MCD), allows users to input a design and a set of desired requirements, and then recommends targeted modifications to the design attributes to achieve these requirements. It identifies these modifications by querying a set of design evaluators in a directed search procedure dictated by an evolutionary algorithm. We demonstrate how predictors ranging from machine learning regressors to text embedding models can support target requirements ranging from functional performance targets to subjective text descriptions. 

MCD can be viewed as an AI-driven design modification tool that allows users to ask challenging objective and subjective questions about an existing design, such as: ``What modifications would it take to make this product 10\% lighter?'', ``What would make my design look like this other concept?'', or ``How would my design need to change to look more sleek and futuristic?'' By enabling designers to interact with AI systems simply and intuitively, counterfactuals open the doors to more successful human-AI collaboration by enhancing and accelerating the design process.

\begin{figure}
    \centering
    \includegraphics[width=0.49\textwidth]{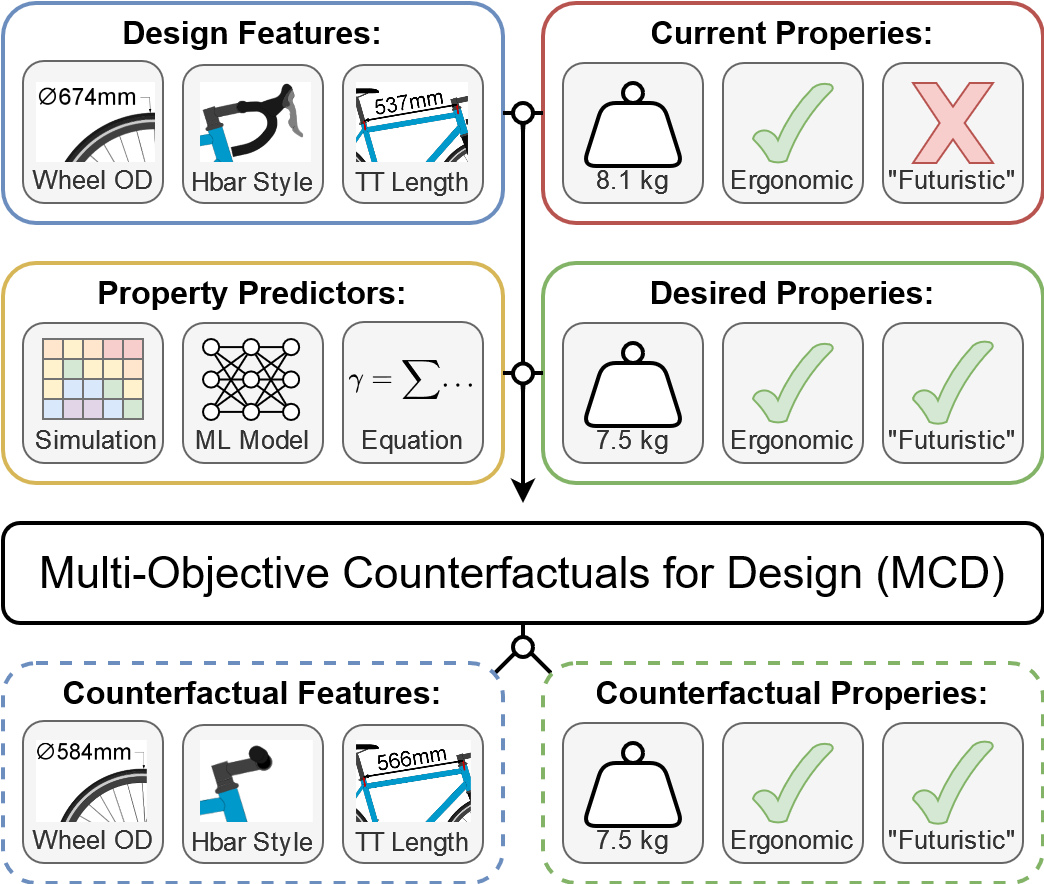}
    \caption{Multi-objective Counterfactuals for Design answers ``Inverse Counterfactual'' design questions --- Given an existing design and its properties, as well as a set of target properties (requirements) and property predictors (design evaluators), MCD identifies variants of the design that achieve the requirements. This enables a human-AI collaborative workflow in which MCD recommends efficient design modifications to achieve designer-specified goals.}
    \label{fig:block}
\end{figure}

In this paper, we showcase our MCD method and demonstrate that counterfactual search is a promising AI-driven design modification tool that designers can leverage for a variety of tasks. To do so, we make several key contributions, which we summarize below. 
\begin{enumerate}
    \item We introduce Multi-objective Counterfactuals for Design (MCD), a new method to search for counterfactual design modifications that achieve desired outcomes. We formulate MCD as a multi-objective search problem to minimize the magnitude and extent of the modifications, encourage proximity to the data manifold, and satisfy user-provided multi-modal requirements.
    \item We demonstrate that MCD effectively suggests targeted design modifications to improve the functional performance of query designs, illustrating that counterfactual search could be viewed as an effective design modification tool. 
    \item We present a text and image-based counterfactual search in design using Contrastive Language-Image Pre-training (CLIP), the first instance of a multi-modal counterfactual search, to our knowledge. In doing so, we demonstrate that counterfactual search can accommodate highly subjective and multi-modal design requirements.  
    \item We demonstrate that MCD can not only tackle both quantitative and subjective requirements, but can also handle highly-specialized requirements like ergonomic fit, which are customized for an individual user. Moreover, we show that MCD can tackle these many multidsciplinary requirements simultaneously and effectively. 
\end{enumerate}

\section{Background}
Counterfactuals are a useful tool for investigating causality and forecasting the potential outcomes of different actions. 
Counterfactuals have been extensively used in various fields, including psychology, philosophy, social sciences, and machine learning. In psychology, counterfactual thinking has been studied in relation to emotions, such as regret and disappointment~\cite{roese2014might}. In philosophy, counterfactuals have been used to explore questions of determinism and free will~\cite{hoerl2011understanding}. In social sciences, counterfactual analysis is widely used to evaluate the impact of policies and interventions~\cite{fearon1991counterfactuals}. Counterfactual explanations are also gaining traction in the field of machine learning as a means to improve the interpretability and fairness of machine learning models~\cite{verma2020counterfactual, guidotti2022counterfactual}.

In this literature review, we discuss several key areas that relate closely to our work --- 1) Optimization in engineering design 2) explainability and counterfactuals in machine learning, 3) design modifications as counterfactuals, 4) multi-objective optimization approaches to counterfactuals, and 5) multi-modal counterfactual search and optimization. 
\subsection{Optimization in Engineering Design}
Optimization is a powerful problem-solving tool in design with a rich theoretical foundation~\cite{papalambros2000principles, gero2012design}. Typically, optimization attempts to find one or many `best' (optimal) designs,  subject to a set of constraints. Optimality is typically evaluated with respect to one or more `objectives.' When multiple objectives are present, it is typical that a single design will not achieve optimal performance across all objectives. Instead of finding just a single design, a non-dominated set can be identified where no design in the set is strictly superior to another in every objective. To effectively identify such non-dominated sets, a variety of effective methods have been proposed in the field of multi-objective optimization (MOO)~\cite{gunantara2018review}. In our case studies, we will use a popular MOO algorithm --- the Nondominated Sorting Genetic Algorithm-II (NSGA-II)~\cite{deb2002fast}. NSGA-II boasts several innovative features, such as non-dominated sorting for elitist selection, crowding distance to encourage diversity, and genetic operators such as tournament selection, simulated binary crossover, and polynomial mutation. 
Our methodology, while demonstrated using NSGA-II, is not restricted to this specific MOO algorithm. The core insights and applicability of our approach remain valid across various MOO algorithms, ensuring broad relevance and adaptability in diverse optimization contexts.

\paragraph{Counterfactual Search Versus Conventional Optimization:}
Though MCD leverages optimization algorithms to generate counterfactuals, it's important to underscore the distinct nature of counterfactual search compared to traditional optimization approaches. Predominantly, conventional optimization endeavors to pinpoint high-performing designs that satisfy a series of objectives. In contrast, counterfactual search is dedicated to finding \emph{minimal modifications} to pre-existing designs to meet these objectives. This distinction enables counterfactual search to thrive in scenarios where explicit constraints might not be delineated, a capability we illustrate in Sec.~\ref{sec:ablation}.

Furthermore, traditional optimization methods often are not data-driven and do not prioritize staying close to the data manifold. MCD, on the other hand, incorporates a data-prior, aiming to ensure that the suggested modifications remain within the bounds of the data manifold. This approach not only fosters more realistic and implementable solutions but also enhances the relevance and applicability of the generated counterfactuals in real-world settings.

\subsection{Counterfactual Explainations in ML}\label{Sec:CFBackground}
A particularly related body of research to our work is counterfactual explanations. Counterfactual explanations were originally developed as a tool to interpret machine learning (ML) models~\cite{wachter2017counterfactual}. Counterfactuals enable users to better understand their models by asking questions about counterfactual scenarios and have become a staple in machine learning explainability. A classic motivating example for counterfactual explanations involves a model that is deciding whether to approve a loan, where the applicant may ask: ``What would I need to change for this model to approve my application?'' Broadly speaking, these counterfactuals answer a very versatile question: ``Hypothetically, what would I need to change about the input to my model for it to predict another outcome?'' Many of the common challenges that designers face can be framed as a similar question. For example, given an evaluator that calculates the functional performance of a design, a designer can ask how to modify the design to achieve some desired functional performance. Despite this, counterfactual explanations have not yet been used in design engineering problems, to the best of our knowledge\footnote{A search for the term ``counterfactual explanations'' on the entire ASME digital collection, that includes design venues such as the IDETC conference and the Journal of Mechanical Design, returned zero results on March 10, 2023.}. 

Since their introduction, researchers have proposed a wealth of counterfactual explanation approaches, which Verma~\etal~\cite{verma2020counterfactual} and Guidotti~\etal~\cite{guidotti2022counterfactual} review. Among the popular methods are Diverse Counterfactual Explanations (DiCE)~\cite{mothilal2020explaining}, Feasible and Actionable Counterfactual Explanations (FACE)~\cite{poyiadzi2020face}, and Multi-Objective Counterfactuals (MOC)~\cite{dandl2020multi}. Counterfactual explanations are a much-needed alternative to popular `feature importance' methods~\cite{ribeiro2016should, lundberg2017unified} in machine learning explainability. In the design automation community, these `feature importance' methods are often used to determine which design parameters have outsized impacts on design performance~\cite{joung2021approach, regenwetter2022biked, rodriguez2020interpretation, dachowicz2022mission, song2023multi} or which parameters are important for relationships between products~\cite{10.1115/1.4054299}. However, like counterfactual explanations, counterfactual-based feature importance has also not been employed in design.
% Another common approach to explainability involves visualizing a model's decisions in some way. This technique lends itself well to data modalities that are easily appreciated visually, such as images, for which saliency maps are a common explainability method~\cite{simonyan2013deep, zeiler2014visualizing, zhou2018interpretable}. 

\subsection{Design Modifications as Counterfactuals}
Counterfactuals are a popular ML explainability tool since they allow users to intuitively understand the ML model's internal decision thresholds (i.e. ``Where does my model start predicting a different outcome?''). Designers and engineers ask many similar questions about their computational tools (i.e. ``For what design variant will my finite element simulation calculate a safety factor greater than 1?''). While ML practitioners leverage counterfactuals to navigate ML models, designers can similarly leverage counterfactuals to navigate both ML models and other design evaluators (numerical simulations, geometry engines, analytical equations, etc.). Even with the abstraction of ML models to general design `evaluators,' many of the guiding principles that have been proposed in ML literature for counterfactual explainations~\cite{verma2020counterfactual, guidotti2022counterfactual} are similarly relevant for design modifications. In general, good design modifications and good counterfactual explanations should share many of the same properties. These include:
% Much like LIME and SHAP, many counterfactuals take a localized approach, with some even fitting local approximations to the data manifold to guide their explanations~\cite{guidotti2019factual}.
% Counterfactuals have most commonly been proposed for tabular data, but have also been applied to images~\cite{goyal2019counterfactual} and text~\cite{hendricks2018generating}, among other modalities. 
\begin{enumerate}
    \item \textbf{Validity of Design Modifications:} First and foremost, a good design modification should result in a desired outcome. For example, if we are querying a geometry engine that calculates the mass of a design and we specify a range of 2-3~kg, a modification should result in a calculated mass in this range. Depending on the nature of the problem, this desired outcome may be a class, an inequality, a range, an exact equality, or some combination of the above. 
    \item \textbf{Sparsity in Number of Modifications:} Good modifications should be easy to realize, meaning that they should not change many features of the query. Sparsity refers to the number of features that must be adjusted to realize a modification. 
    \item \textbf{Proximity and Minimal Adjustments:} While the number of adjustments needed to realize a modification is an important consideration, the extent of these adjustments is also important. In simple terms, we would like the modified design to be as similar to the query design as possible. This is typically quantified as a distance to the original query. 
    \item \textbf{Manifold Proximity for Realism and Validity:} When modifications are guided by statistical evaluators, their accuracy is contingent upon the evaluator's familiarity with its training data. When modifications venture too far from the data manifold---essentially the space of known, validated designs on which the evaluator was trained---their predicted outcomes become unreliable. This challenge underscores the importance of manifold proximity, which serves a dual purpose: it not only ensures the evaluator's predictions remain accurate but also boosts the realism and validity of the proposed design changes. In essence, designs that closely resemble or stay within the manifold of existing, proven designs are inherently more likely to be both feasible and effective. This principle leverages the underlying data to implicitly enforce constraints, guiding towards solutions that are not just optimal but practically viable as well.
    % \item \textbf{Manifold Proximity:} When a modification is based on the estimate of a statistical evaluator, the evaluator's accuracy will depend on the support of its training dataset. If queries lie too far from the data manifold on which the evaluator was trained, predictions for potential modifications will no longer be accurate. However, even when evaluators are deterministic, manifold proximity can help implicitly enforce constraints. Simply put, a design is more likely to be valid and viable if it is similar to existing known designs. 
    \item \textbf{Actionability of Design Modifications:} In many problems, certain input parameters may not be changeable, but will nonetheless play a role in the output of the model. For example, the weight of the rider will play a significant role in the structural loading of a bicycle. However, when designing a bicycle, we can't choose to simply make the rider lighter. A good design modification should only modify actionable features. 
    % Several works, such as~\cite{poyiadzi2020face}, have also proposed more nuanced methods to handle actionability. 
    \item \textbf{Causality and Design Coherence:} Design features are often causally linked, implying that changing one feature may necessitate changing another. 
    For example, increasing the diameter of a rotary shaft will typically require an increase in the hole diameter of any components that interface with that shaft, such as bearings. Recognizing and adhering to these causal links is key in devising modifications that are not just theoretically sound but also practically viable. Effective design modifications should thus embody a deep understanding of these causal dynamics, ensuring that changes are coherent and aligned with the underlying physical principles governing the design.

\end{enumerate}
There is ample reason to treat design modifications as counterfactuals. Naturally, a strong counterfactual generator will identify modifications with the above properties. However, good counterfactual generators should also exhibit several properties that may not be reflected in the strength of individual modifications themselves:
\begin{enumerate}
    \item \textbf{Diverse Sets of Modifications:} As emphasized in~\cite{mothilal2020explaining}, it may be highly desirable to generate diverse sets of modifications. This gives the user a wealth of options, ideally with different actionable requirements to achieve the query objective.
    \item \textbf{Model-Agnosticism in Modification Systems:} Ideally, the modification system should treat the evaluator as a ``black box'' and require nothing more than a simple evaluation function~\cite{verma2020counterfactual}. These ``model-agnostic'' methods allow for wider applicability and code reuse. Notably, model-agnostic approaches do not rely on gradient information from the evaluator but may be less sample-efficient than methods that leverage gradients, when available. 
\end{enumerate}

\subsection{Multi-Objective Counterfactual Explanations}
As discussed, counterfactual search is an effective method to generate design modifications. Counterfactual search can be viewed as an optimization problem, and can similarly be implemented using an optimization algorithm. Many methods summarize the optimization objective as a weighted sum of the different objectives discussed earlier. However, another approach instead frames the counterfactual search process as a classic multi-objective optimization (MOO) problem. Dandl~\etal~\cite{dandl2020multi} formalize this parallel between counterfactual search and MOO in Multi-Objective Counterfactuals (MOC). By handling objectives individually rather than as a single aggregated objective, MOC realizes a key benefit of Multi-Objective Optimization, namely the ability to generate non-dominated sets of counterfactuals. Whereas a single-objective approach returns a counterfactual that optimizes for a statically weighted aggregation of objectives, the non-dominated set allows designers to adaptively select counterfactuals based on their specific search priorities, which typically depend on the problem at hand.

Though MOC formalized counterfactual search as an optimization problem, its principal concern was the use of counterfactuals for machine learning explainability. Though similar in a few aspects, MCD expands upon (MOC) in several key directions focused on applicability to design modification. Chiefly, despite its name, MOC does not inherently support multi-objective queries. Furthermore, MOC does not distinguish between hard and soft constraints, despite the fact that this functionality is ingrained in the Non-Dominated Sorting Genetic Algorithm II (NSGA-II)~\cite{deb2002fast} that MOC is built around. MCD addresses these gaps while also decoupling the optimization and sampling steps, and introducing new ways to integrate counterfactuals with representation-learning models to enable multi-modal counterfactual search. 

Since the overarching goal of MCD is not to explain evaluators, but rather to search for counterfactuals that can be used as design modifications, we refer to the problem as `counterfactual search.' Unlike counterfactual explanations, counterfactual search does not require ML predictors and can work with many types of evaluators. It also has the additional goals of manifold similarity and meeting multi-objective multi-modal requirements.

% Cite other interpretability papers.

\subsection{Multi-Modal Counterfactuals}~\label{crossmodal}
The multitude of data modalities (parametric, images, meshes, etc.) spanned by design data remains a prominent challenge in data-driven design~\cite{regenwetter2022deep, song2023multi, regenwetter2023beyond}. Existing inverse design tools, particularly optimization frameworks, seldom support design targets specified using complex data modalities. For example, interacting with design optimization tools using text prompts is practically unheard of, but may be very intuitive and desirable. In this paper, we demonstrate that MCD (and conventional optimization algorithms) can be enhanced with representation-learning AI models to process design requirements in a variety of multi-modal data representations. However, setting appropriate design constraints in multi-modal settings can be challenging. As we showcase in Sec.~\ref{sec:ablation}, MCD's ability to handle implicit constraints makes it more robust than conventional optimization algorithms in underconstrained multi-modal problems.

% Though a model explained by a counterfactual method may make predictions in one modality, users may instead prefer to query targets in an entirely different modality. This 

% We will demonstrate in this paper that MCD can be used in conjunction with rendering pipelines and trained language models to generate counterfactuals for a parametric model using images or even text prompts. In this way, counterfactuals can capture complex and abstract user requirements in a `zero-shot' fashion, requiring no additional training to understand the context of the prompts. To provide context for this discussion, we will introduce a brief background on relevant subjects in cross-modal learning. 

When handling data of modalities like graphs~\cite{cai2018comprehensive}, images~\cite{faghri2017vse++}, 3D geometry~\cite{dai2018siamese}, text~\cite{devlin2018bert, cer2018universal}, and mixed modalities, a common representation-learning technique involves mapping datapoints to a vector space. This effectively creates a link from datapoints of the modality to datapoints in the vector space. Two or more modalities can then be linked by creating shared embeddings for the modalities using the same vector space. A prominent example of this is Contrastive Language-Image Pretraining (CLIP)~\cite{radford2021learning}, a framework in which text and images are mapped to a shared embedding space. CLIP models are rewarded for mapping matching pairs to similar embedding vectors and mapping non-matching pairs to dissimilar embedding vectors. CLIP has garnered significant attention in recent years by enabling high-performing vision-language models~\cite{rombach2022high}. In our experiments, we will be leveraging pre-trained CLIP models to query counterfactuals using text and image prompts. 

\section{Methodology}
In this section, we discuss the optimization formulation behind MCD, emphasizing the constraints, objectives, and operators used. We then present our approach for sampling diverse sets of counterfactuals and discuss how we decouple the optimization from the final sampling step. Finally, we demonstrate the capabilities of MCD on a simple 2D problem.

\subsection{Objectives for Quality Design Modifications}
Broadly, MCD considers two types of objectives: Objectives related to counterfactual quality and user-specified auxiliary objectives (often used for soft constraints). The former draws on MOC~\cite{dandl2020multi}, which leverages Gower distance~\cite{gower1971general} and the number of changed features as optimization objectives in MOC. In the following objectives, we consider a query (original design), $q$, and a counterfactual (modified design), $x$, which are each represented as a $d$-dimensional design vector. We also assume that exactly $n$ evaluation functions are given by the user, which are used to calculate performance objectives, constraint satisfaction, or both.  
\begin{enumerate}
    \item \textbf{Proximity to Query:} To calculate proximity, we use Gower Distance~\cite{gower1971general}, a metric that indicates the distance between mixed-type data points. The Gower distance between d-dimensional counterfactual $x$ and query $q$ is given in terms of their feature values $x_i$ and $q_i$ for $i\in[1...d]$, as:
    \begin{equation}
        f_{pr}(x,q) = \frac{1}{d}\sum_{i=1}^d\delta_G(x_i,q_i)
    \end{equation}
    $\delta_G(x_i,q_i)$ is a function that depends on feature type and is given as: 
    \begin{equation}
        \delta_G(x,q) = 
        \begin{cases} 
        \frac{1}{\hat R_i}|x_i-q_i| & \text{if }x_i\text{ is numerical} \\
        \mathbbm{1}_{x_i\neq q_i} & \text{if }x_i\text{ is categorical}
        \end{cases}
    \end{equation}
    Here, $\hat R_i$ is the range of the feature $i$ observed in the dataset. 
    \item \textbf{Sparsity of Modification:} To calculate sparsity, we quantify the proportion of features that the counterfactual, $x$, modifies from the query, $q$. 
    \begin{equation}
        f_{sp}(x,q) = \frac{||x-q||_0}{d} = \frac{1}{d}\sum_{i=1}^d\mathbbm{1}_{x_i\neq q_i}
    \end{equation}
    \item \textbf{Proximity to Dataset Manifold:} To measure the manifold proximity, we calculate the average Gower distance to the $k$ nearest observed data points $s^i...s^k$ from the dataset $S$, where k is a tuning parameter:
    \begin{equation}
        f_{mp}(x,S) = \frac{1}{k}\sum_{i=1}^k\frac{1}{d}\sum_{j=1}^d\delta_G(x_j,s_j^{i})
    \end{equation}
    \item \textbf{Problem-Specific Objectives:} Users may also specify objectives pertaining to the output of the evaluators $f_i(x)\,\,\,\forall\,\,\, i\in\mathbb{O}$, where $\mathbb{O}$ is the set of evaluation functions to be treated as objectives. These auxiliary objectives are directly included as objectives in the optimizer. As we discuss next in \ref{sec:constraints}, users can also choose to place hard constraints on the outputs of evaluators, allowing them to serve as either performance evaluators, constraint evaluators, or both. 
\end{enumerate}

\subsection{Constraints on Valid Design Modifications} \label{sec:constraints}
In a counterfactual search, a variety of optimization constraints may be present. Constraints are considered non-negotiable and always take precedence over objectives. 
In practice, many optimization algorithms, including MCD's underlying NSGA-II optimizer, prioritize resolving constraint violations before proceeding to the optimization of objectives. MCD considers several types of constraints:
\begin{enumerate}
    \item \textbf{Evaluator Output Constraints:} 
    Users may desire to apply arbitrary constraints, as is classic in multi-objective optimization. Users can do so by affixing inequality constraints to the output of evaluators, effectively making them constraint evaluation functions. Evaluator output constraints are also the classic motivating case for counterfactual explanations in ML. To maintain generality, we represent these constraints with inequalities using lower bounds $L_i$ and upper bounds $U_i$ for evaluators $f_i(x)\,\, \forall \,\, i\in\mathbb{C}$, where $\mathbb{C}$ is the set of evaluation functions to be treated as constraints. Note that $\mathbb{C}$ and $\mathbb{O}$ may overlap. The intersection, $\mathbb{C}\cap\mathbb{O}$, represents the set of evaluation functions that are both constrained and used as an objective (often representing performance objectives with a hard requirement). We write this constraint as: 
    \begin{equation}
        L_j\leq f_j(x)\leq U_j\,\,\forall\,\,j \in {\mathbb{C}}
    \end{equation}

    \item \textbf{Parameter Bounds:} MCD also allows users to specify bounds for continuous design parameters or limited sets that categorical variables can select from. We define the lower bound $l_k$ and upper bound $u_k$ for parameter $k$ where $x_k$ is continuous or discrete and ordered. If $x_k$ is instead categorical, we define a set of valid parameter values $\mathcal{V}_k$ for parameter $k$. For $k\in\{1...d\}$, where d is the dimensionality of the design space: 
    \begin{flalign}
    \begin{aligned}
        l_k\leq x_k\leq u_k\,\,\,& \text{if }x_k\text{ is numerical} \\
        x_k \in \mathcal{V}_k\,\,\,& \text{if }x_k\text{ is categorical}
    \end{aligned}
    \end{flalign}
    
    \item \textbf{Actionable Features:} Like many counterfactual models, we implement a mechanism to constrain which features are allowed to be modified by a counterfactual, as specified by the user. We call the set of actionable features $\mathcal{A}$. Any feature $l\in\{1...d\}$ that is not actionable must not deviate from the query: 
    \begin{equation}
    x_l=q_l, \,\,\, \text{if }l \not\in \mathcal{A} 
    \end{equation}

\end{enumerate}
\subsection{Formulation as AI-driven MOO problem} \label{sec:MOOformulation}
In summary, we express the multi-objective optimization problem in terms of the variables, sets, and functions defined above as follows, where $x^*$ is an optimal choice of $x$:
\begin{flalign}
\begin{aligned}
    \text{minimize: } &f_i(x^*),\,\,\forall\,\,i\in \{pr, sp, mp\}\cup\mathbb{O} \\
    \text{subject to: } &f_j(x^*)-L_j\geq 0, \\
    &U_j - f_j(x^*)\geq 0, \\
    &u_k - x_k^*\geq 0, \,\,\, \text{if }x_k\text{ is numerical}, \\
    &x_k^*-l_k\geq 0, \,\,\, \text{if }x_k\text{ is numerical}, \\
    &x^*_k \in \mathcal{V}_k,\,\,\, \text{if }x_k\text{ is categorical},\\
    &x^*_l=q_l, \,\,\, \text{if }l \not\in \mathcal{A},\\
    &\forall\,\,j\in \mathbb{C}, \,\,\, k \in \{1...d\},\,\,\, l \in \{1...d\}
\end{aligned}
\end{flalign}

\subsection{Algorithm}
Any gradient- or non-gradient-based multi-objective optimization method could be used to optimize MCD problems. We implement a backend using the Non-Dominated Sorting Genetic Algorithm II (NSGA-II)~\cite{deb2002fast} to generate the results in the paper. Notably, NSGA-II is a model-agnostic optimizer, requiring no gradient information from the predictors, allowing for the widest applicability across design problems. We use an implementation of NSGA-II from~\cite{pymoo}, including the mixed-variable selection, crossover, and mutation functions provided. 

The initial population always consists of the query and a set of randomly sampled points from the confines of the user-specified design space or directly from the dataset. In problems with continuous variables, we find that without any precautions to maintain the exact parameter values from the original query, these values tend to get `lost,' and can never be exactly reconstructed, hurting the sparsity objective of counterfactuals. To allow the algorithm to `rediscover' the exact parameter values from the query, we introduce a custom `repair' operator that randomly reverts individual parameter values back to the query's values with a certain probability.

\subsection{Sampling}
Contrary to other counterfactual search approaches, MCD decouples the optimization and sampling steps. Conventionally, a user will have to decide on the priorities between various objectives (e.g. proximity, diversity, manifold proximity, etc.) before running the optimization. This is impractical, as these objectives are challenging to select intuitively, and must often be chosen through trial and error. For example, a designer might realize that the generated counterfactuals are much too different from the query to be practically realizable. By avoiding retraining, our method can save significant computational expense and, as we will discuss in Sec.~\ref{CS2}, enable users to quickly consider counterfactuals from different regions of the objective landscape. 
We decouple the search and sampling process as follows: 
\begin{enumerate}
    \item Given a query, a set of constraints, and objectives, the optimizer generates a collection of candidate counterfactuals by running NSGA-II. 
    \item The sampling algorithm collects a set of objective priority weights from the user. By collecting these weights after training, MCD allows rapid counterfactual sampling under different objective weights without the need for retraining, unlike other approaches.  
    \item Each candidate counterfactual is assigned an aggregate quality score, which is calculated as a sum of individual objective scores, weighted by their priority. Objectives can also be provided with specified targets, in which case the Design Target Achievement Index~\cite{regenwetter2022design} is used to quantify target achievement before factoring into the aggregate score. The aggregate score, $f_{a}$ of a counterfactual candidate, x, is given in terms of objective priority weights $w_{pr}, w_{sp}, w_{mp}$ by:
    \begin{flalign}
    \begin{aligned}
        f_{a}(p) = &w_{pr}f_{pr}(x,q)\\
        +\,&w_{sp}f_{sp}(x,q)\\
        +\,&w_{mp}f_{mp}(x,S)\\
        +\,&\sum_{i\in\mathbb{O}}w_if_i(x)
    \end{aligned}
    \end{flalign}
    Here, $w_i$ are weights for the user-specified auxiliary objectives, $i\in\mathcal{O}$.  
    \item A performance-weighted diversity matrix is calculated using a Gower distance-based similarity kernel, $\delta_G(i,j)$ to evaluate the similarity between counterfactuals. Matrix entries are calculated as a function of aggregate scores and a diversity parameter, $w_d$ as:
    \begin{equation}
        D_{i,j}=\delta_G(i,j)\left(f_{a}(i)f_{a}(j)\right)^\frac{1}{w_d}
    \end{equation}
    \item A diverse subset of high-performing counterfactuals is selected from this matrix using greedy diverse subset selection. 
\end{enumerate}

\subsection{Requirements}
A block diagram demonstrating MCD's anticipated usage scenario is shown in Figure~\ref{fig:block}. MCD's requirements can be summarized as follows:
\begin{enumerate}
    \item \textbf{Query} (required): MCD needs an original design, which is referenced in proximity and sparsity calculations. 
    \item \textbf{Evaluators} (required): MCD needs one or several evaluators to calculate the performance of design candidates. These may be numerical simulations, analytical equations (as in Sec.~\ref{sec:2d}), predictive surrogate models (as in Sec.~\ref{CS1}), an entire evaluation pipeline (as in Sec.~\ref{CS2}), or something else. 
    \item \textbf{Evaluator Output Constraints} (required): Users must define at least one constraint on the output of an evaluator. This is the counterfactual outcome that the user wants to achieve and is a necessity for counterfactual search.
    \item \textbf{Dataset} (recommended): MCD's manifold proximity is calculated by referencing a dataset. However, the dataset can optionally be foregone and the manifold proximity unused, though several features of MCD will be lost. 
    \item \textbf{Design Space Constraints} (optional): As discussed in Sec.~\ref{sec:constraints}, the user can constrain actionable features and parameter bounds to constraint the design space over which MCD can explore
    \item \textbf{Auxiliary Objectives} (optional): The user may optionally specify auxiliary objectives or performance targets for MCD to consider. 
    
\end{enumerate}

\subsection{Showcasing Functionality on 2D Examples} \label{sec:2d}
\begin{figure}
     \captionsetup[subfigure]{justification=centering}
     \centering
     \captionsetup[subfigure]{aboveskip=-1pt,}
     \begin{subfigure}[b]{0.24\textwidth}
         \centering
         \includegraphics[width=\textwidth]{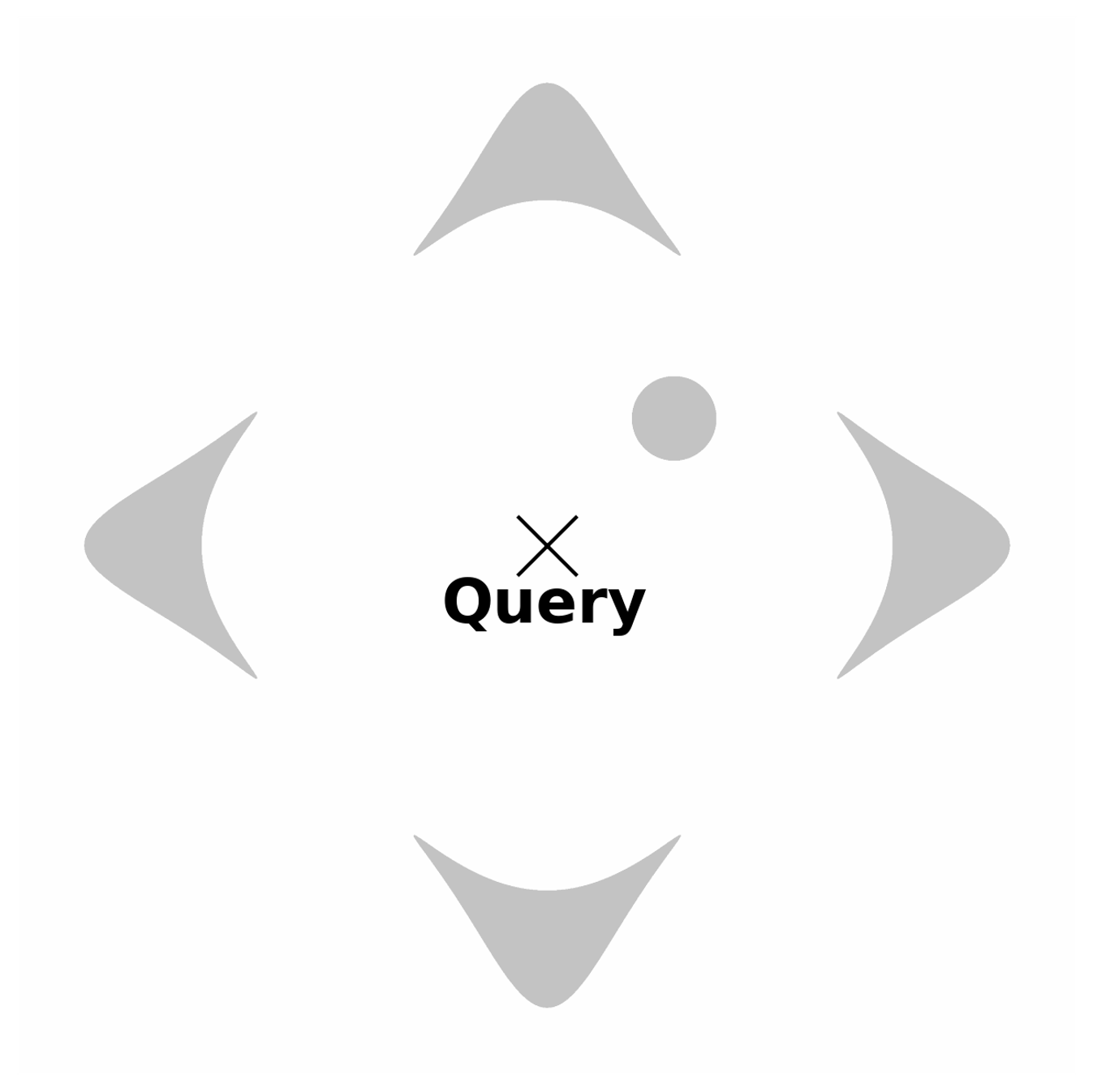}
         \caption{Problem setup (target region shown in gray)}
         \label{fig:2Dregions}
     \end{subfigure}
     \begin{subfigure}[b]{0.24\textwidth}
         \centering
         \includegraphics[width=\textwidth]{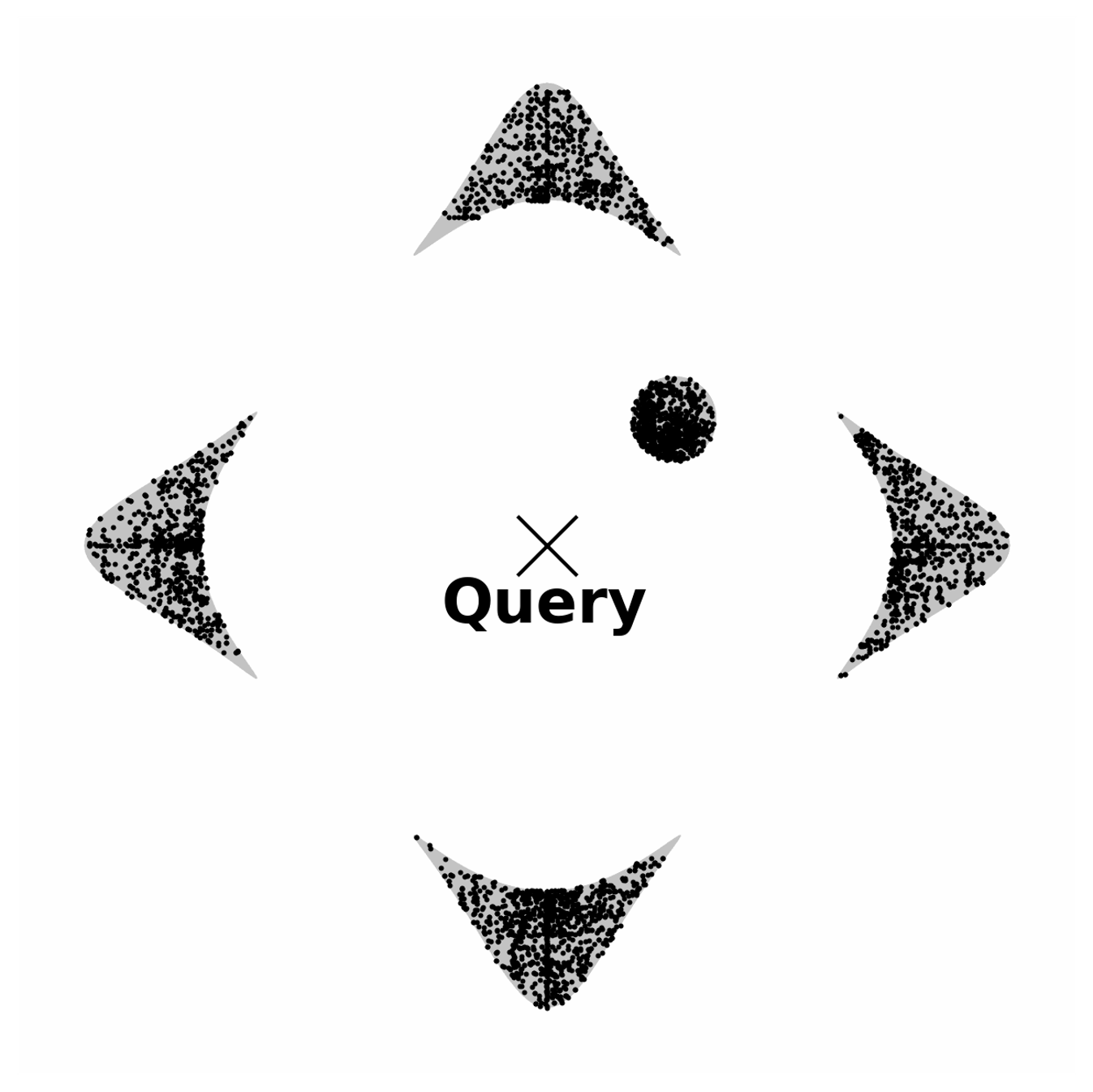}
         \caption{All solutions found during optimization}
         \label{fig:2Dall}
     \end{subfigure}

     \begin{subfigure}[b]{0.24\textwidth}
         \centering
         \includegraphics[width=\textwidth]{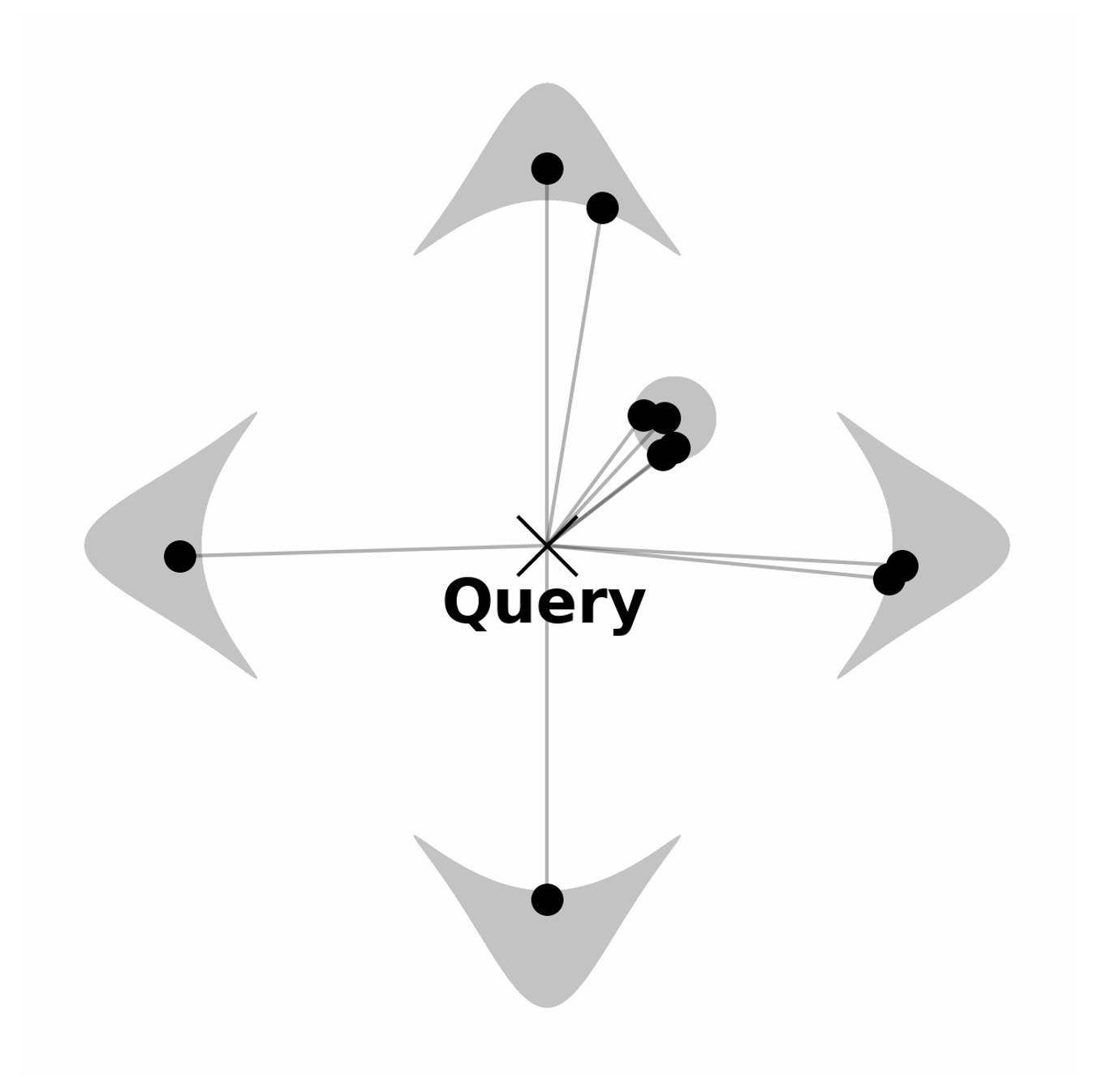}
         \caption{Sampling using balanced objective weights}
         \label{fig:2Dbalanced}
     \end{subfigure}
     \begin{subfigure}[b]{0.24\textwidth}
         \centering
         \includegraphics[width=\textwidth]{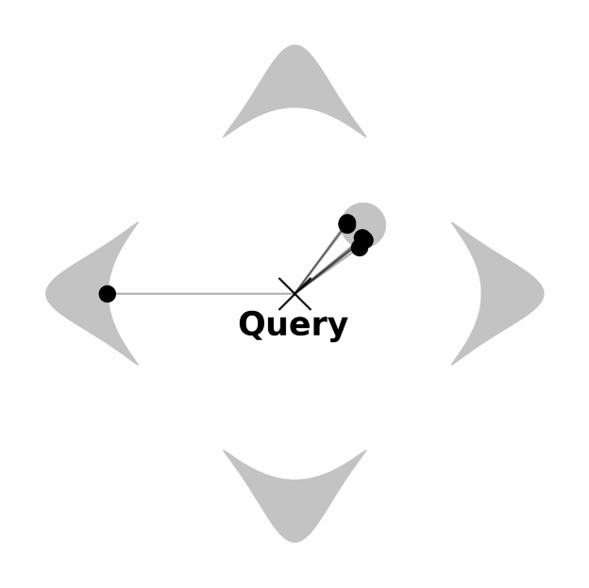}
         \caption{Sampling using high proximity weight}
         \label{fig:2Dproximity}
     \end{subfigure}

     \begin{subfigure}[b]{0.24\textwidth}
         \centering
         \includegraphics[width=\textwidth]{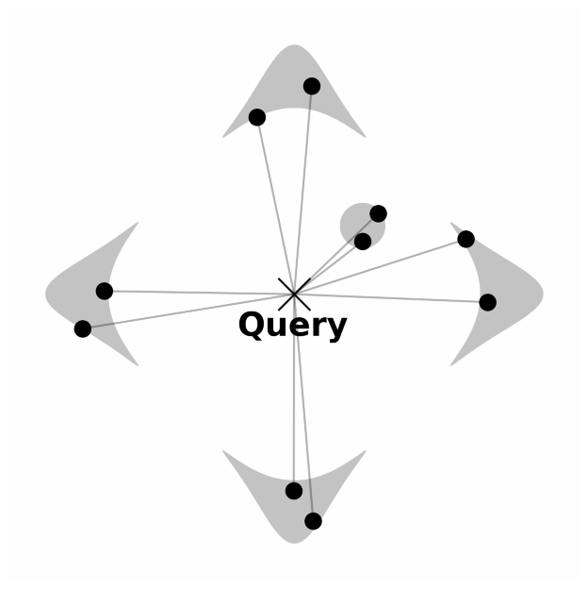}
         \caption{Sampling using high \\diversity weight}
         \label{fig:2Ddiverse}
     \end{subfigure}
     \begin{subfigure}[b]{0.24\textwidth}
         \centering
         \includegraphics[width=\textwidth]{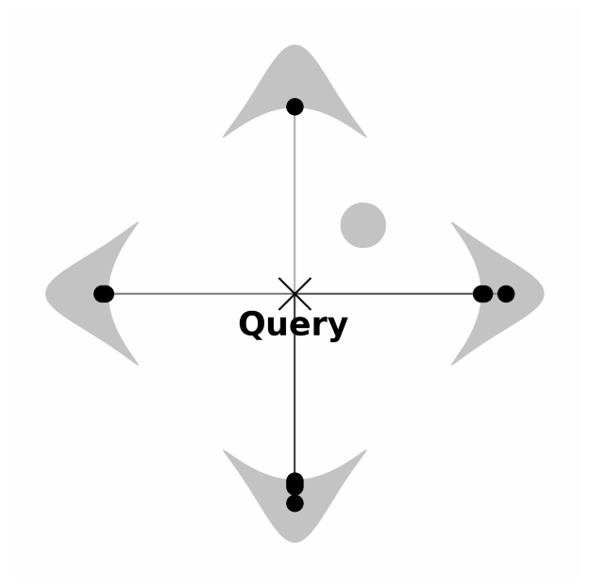}
         \caption{Sampling using high \\sparsity weight}
         \label{fig:2Dsparse}
     \end{subfigure}

    \caption{Demonstration of inverse counterfactual search in 2D space. The relative priority weighting of objectives has a significant impact on counterfactuals sampled during sampling.}
    \label{fig:2D}
\end{figure}
% \begin{figure}
%      \centering
%      \captionsetup[subfigure]{aboveskip=-1pt,}
%      \begin{subfigure}[b]{0.49\textwidth}
%          \centering
%          \includegraphics[width=\textwidth]{Images/2dbalanced.png}
%          \caption{Balanced Sampling ($w_{pr} = 0.5,\,w_{sp}=0.2,\,w_{mp}=0.5,\,w_d=0.2$)}
%          \label{fig:2Dbalanced}
%      \end{subfigure}

%      \begin{subfigure}[b]{0.49\textwidth}
%          \centering
%          \includegraphics[width=\textwidth]{Images/2dproximity.png}
%          \caption{High Prox. Weight ($w_{pr} = 50,\,w_{sp}=0.2,\,w_{mp}=0.5,\,w_d=0.2$)}
%          \label{fig:2Dproximity}
%      \end{subfigure}

%      \begin{subfigure}[b]{0.49\textwidth}
%          \centering
%          \includegraphics[width=\textwidth]{Images/2ddiverse.png}
%          \caption{High Diversity Weight ($w_{pr} = 0.5,\,w_{sp}=0.2,\,w_{mp}=0.5,\,w_d=20$)}
%          \label{fig:2Ddiverse}
%      \end{subfigure}

%      \begin{subfigure}[b]{0.49\textwidth}
%          \centering
%          \includegraphics[width=\textwidth]{Images/2dsparse.png}
%          \caption{High Sparsity Weight ($w_{pr} = 0.5,\,w_{sp}=20,\,w_{mp}=0.5,\,w_d=0.2$)}
%          \label{fig:2Dsparse}
%      \end{subfigure}

%     \caption{Counterfactual sets returned for three query designs under different weightings of counterfactual quality objectives. Performance space constraints are indicated on the plots. Valid counterfactuals must simultaneously meet both constraints.}
%     \label{fig:2D}
% \end{figure}

Before showcasing the capabilities of MCD on real design datasets, we will first demonstrate its functionality on a simple two-dimensional problem for ease of visualization. In this problem, five distinct regions of a two-dimensional design problem are designated valid, and we have a starting query that is in none of the regions (Fig. \ref{fig:2Dregions}). We seek a design modification that will be valid. We first optimize for several generations, yielding thousands of possible solutions (Fig. \ref{fig:2Dall}), then sample high-performing counterfactual sets of 10 designs from the set of all valid solutions identified during optimization. During sampling, we consider different choices of objective weights ($w_{pr}$, $w_{sp}$, $w_{mp}$, $w_d$). 
\begin{enumerate}
    \item When sampling using ``balanced'' objective weights, the sampled counterfactual sets achieve a balance of proximity, diversity, and sparsity. Designs in the most proximal region are favored, but at least one design is sampled from each region (Fig.~\ref{fig:2Dbalanced}). 
    \item When proximity weight ($w_{pr}$) is relatively high, most counterfactuals in each set are sampled from the mode nearest to the query (Fig.~\ref{fig:2Dproximity}).
    \item When diversity weight ($w_{d}$) is relatively high, two counterfactuals are sampled from each mode (Fig.~\ref{fig:2Ddiverse}).
    \item Finally, when sparsity ($w_{sp}$) is relatively high, counterfactuals are only sampled from locations in the space where only one parameter differs from the query (Fig.~\ref{fig:2Ddiverse}).
\end{enumerate}
Each of these subsets is sampled from the same set of counterfactual candidates with no re-optimization necessary. Now, having demonstrated MCD's functionality on a simple 2D problem, we move on to a more complex real-world design problem: Bicycle design. The following experiments will utilize the BIKED ecosystem~\cite{regenwetter2022biked, regenwetter2023framed, regenwetter2024biked}, a set of datasets, predictive models, and other computational tools aimed at facilitating data-driven bicycle design. We provide an overview of BIKED in the appendix.

\section{Modifying Designs to Maximize Structural Efficiency} ~\label{CS1}
In our first case study, we consider the counterfactual: ``What if my design were 30\% lighter?'' Specifically, we consider a bicycle frame design problem where we are trying to improve the structural properties and reduce the weight of a given query design. The query, $q$, and any counterfactual $x$ are represented using a set of 37 continuous and categorical variables (e.g. [Steel, 565.6 ... 1.24, 4.26]). We use a regression model trained on the FRAMED dataset consisting of Finite Element (FE) simulation results from 4500 community-designed bike frames~\cite{regenwetter2023framed}, including weight, safety factors, and deflections under various loading conditions. The trained regression model is an AutoGluon~ tabular AutoML regressor~\cite{erickson2020autogluon} trained to accurately predict various structural performance attributes of bicycle frames. 

To illustrate MCD's capabilities, we define three distinct but related problems. The first has a single objective: finding counterfactuals that reduce the predicted mass of the given design. The second has two competing objectives: Maximize a design's safety factor while minimizing its mass. The third has the same objectives as the second but restricts MCD to only vary a more constrained and actionable set of features. In each example, we query the same design $q$: a steel tube road bike with minor structural inefficiencies. Experts would likely suggest that the bike's insufficiently thick down tube is a key shortcoming of the design, requiring other components to be over-engineered to compensate. This bike has a safety factor\footnote{We use predicted safety factor in FRAMED's in-plane loading scenario~\cite{regenwetter2023framed}.} of 1.24 and a mass of 4.26~kg\footnote{Each optimization ran for 100 generations with a population size of 500.}. 

\paragraph{Single objective query:} In the first variant, MCD is tasked with finding counterfactuals that reduce the mass of the original design from 4.26~kg to under 3~kg. Mathematically, $f_1(x)=R_{mass}(x)$, $U_1=3~\text{kg}$, and $L_1 = -\infty~\text{kg}$. Here, $R_{mass}$ is a regression model trained to estimate the mass of designs. MCD successfully discovers hundreds of valid counterfactuals and samples a set of three diverse counterfactuals, $x^{*1-3}$ which have, on average, a mass of 2.0~kg, as tabulated in Table~\ref{q1}. Although MCD succeeds in its explicitly stated objective, a closer look reveals that it does nothing to remedy the wall thickness issue in the down tube, and as a consequence of weight savings in other parts of the sampled frames, the average safety factor across sampled counterfactuals is an abysmal 0.52. This disregard for secondary objectives is quite characteristic of the many existing single-objective counterfactual search algorithms and illustrates why MCD's novel support of multi-objective queries is so essential for design problems. Our next example showcases how to leverage multi-objective requirements to avoid these issues. 

\begin{table}[!htb]

\setlength{\tabcolsep}{3pt}
\caption{Generated counterfactuals $x^{*1-3}$ (CF) for a \textbf{Single Objective Query}. 34 columns are omitted. Like many single-objective counterfactual engines, MCD tends to achieve single-objective queries at the expense of secondary objectives. MCD's unique support of multi-objective queries remedies this problem.  }
\resizebox{.99\columnwidth}{!}{
\begin{tabular}{@{}lcccccc@{}}

\toprule
      & Material & \begin{tabular}[c]{@{}c@{}}Stack\\ (mm)\end{tabular} & ... & \begin{tabular}[c]{@{}c@{}}Down Tube\\ Thick. (mm)\end{tabular} & \begin{tabular}[c]{@{}c@{}}Safety\\ Factor\end{tabular} & \begin{tabular}[c]{@{}c@{}}Frame\\ Mass (kg)\end{tabular} \\
\midrule
q (Query) & Steel    & 565.6                                                & ... & 0.52                                                               & 1.24                                                    & 4.26                                                      \\
\midrule
$x^{*1}$ (CF)  & Steel    & 570.8                                                & ... & 0.52                                                               & 0.52                                                    & 1.99                                                      \\
$x^{*2}$ (CF)  & Steel    & 565.6                                                & ... & 0.52                                                               & 0.27                                                    & 1.64                                                      \\
$x^{*3}$ (CF)  & Steel    & 565.6                                                & ... & 0.52                                                               & 0.76                                                    & 2.48                                                      \\
% $x^{*4}$ (CF)  & Steel    & 565.6                                                & ...  & 0.52                                                               & 0.64                                                    & 2.69                                                      \\
% $x^{*5}$ (CF)  & Aluminum    & 522.6                                                & ...    & 0.52                                                               & 0.22                                                    & 2.70             \\                                        
\bottomrule
\end{tabular}
}
\label{q1}
\end{table}

\paragraph{Bi-objective query:} In the second variant, a second objective is introduced: Increase the safety factor to a minimum value of 1.5. Mathematically, $f_2(x) = R_{SF}(x)$, $U_2 = \infty$, and $L_2=1.5$, where $R_{SF}(x)$ is a regression model trained to estimate the safety factor of a design, $x$. Again, MCD successfully discovers numerous counterfactuals, and the diverse 3-bike sample set has an average mass of 2.3 kg and a safety factor of 1.8, as shown in Table~\ref{q2}. This time, MCD realizes that the bike can be made significantly more weight-efficient by increasing the down tube wall thickness to relieve structural stress on other components to be lightened. However, it also changes the material of the bike from steel to aluminum or titanium in all counterfactuals, a modification that would likely carry a significant increase to the cost and may thus be unactionable. In the presence of a cost prediction model, MCD could consider cost as another query objective. However, even without such a model, MCD can be ordered to leave certain design parameters unchanged, as we demonstrate in our final example.   
\begin{table}[!htb]
\setlength{\tabcolsep}{3pt}

\caption{Generated counterfactuals $x^{*1-3}$ (CF) for a \textbf{Bi-objective query}. 34 columns are omitted. By querying multiple objectives simultaneously, MCD avoided the safety factor issue that occurred in variant 1.}
\resizebox{.99\columnwidth}{!}{
\begin{tabular}{@{}lcccccc@{}}

\toprule
      & Material & \begin{tabular}[c]{@{}c@{}}Stack\\ (mm)\end{tabular} & ... & \begin{tabular}[c]{@{}c@{}}Down Tube\\ Thick. (mm)\end{tabular} & \begin{tabular}[c]{@{}c@{}}Safety\\ Factor\end{tabular} & \begin{tabular}[c]{@{}c@{}}Frame\\ Mass (kg)\end{tabular} \\
      \midrule
q (Query) & Steel    & 565.6                                                & ... & 0.52                                                            & 1.24                                                    & 4.26                                                      \\
\midrule
$x^{*1}$ (CF)  & Aluminum & 565.0                                                & ... & 2.20                                                            & 1.91                                                    & 2.81                                                      \\
$x^{*2}$ (CF)  & Titanium & 561.6                                                & ... & 2.46                                                            & 1.82                                                    & 2.21                                                      \\
$x^{*3}$ (CF)  & Aluminum & 532.2                                                & ... & 1.81                                                            & 1.58                                                    & 1.75                                                      \\
% $x^{*4}$ (CF)  & Titanium & 563.5                                                & ... & 3.92                                                            & 1.60                                                    & 2.23                                                      \\
% $x^{*5}$ (CF)  & Steel    & 565.6                                                & ... & 2.48                                                            & 1.65                                                    & 2.87             \\                                        
\bottomrule
\end{tabular}
}
\label{q2}
\end{table}

\paragraph{Bi-objective query with constraints} In the third variant, MCD is no longer allowed to vary frame material. Mathematically, we specify that the material parameter (mat) of $x$ and $q$ must be the same: $x_{mat} = q_{mat}= \text{Steel}$. MCD proceeds to find tens of valid designs through variations in certain tube diameters, lengths, and other structural configurations. From these valid designs, a 3-bike sample set achieves an average mass of 2.7~kg and an average safety factor of 1.9, as shown in Table~\ref{q3}.

\begin{table}[!htb]

\setlength{\tabcolsep}{3pt}
\caption{Generated counterfactuals $x^{*1-3}$ (CF) for a \textbf{Bi-Objective query with constraints}. 34 columns are omitted. When restricted from modifying frame material, MCD is still able to recommend design modifications that meet the safety factor and mass targets.}
\resizebox{.99\columnwidth}{!}{
\begin{tabular}{@{}lcccccc@{}}

\toprule
      & Material & \begin{tabular}[c]{@{}c@{}}Stack\\ (mm)\end{tabular} & ... & \begin{tabular}[c]{@{}c@{}}Down Tube\\ Thick. (mm)\end{tabular} & \begin{tabular}[c]{@{}c@{}}Safety\\ Factor\end{tabular} & \begin{tabular}[c]{@{}c@{}}Frame\\ Mass (kg)\end{tabular} \\
\midrule
q (Query) & Steel    & 565.6                                                & ... & 0.52                                                            & 1.24                                                    & 4.26                                                      \\
\midrule
$x^{*1}$ (CF)  & Steel    & 565.6                                                & ... & 2.44                                                            & 2.05                                                    & 2.93                                                      \\
$x^{*2}$ (CF)  & Steel    & 601.7                                                & ... & 3.38                                                            & 2.06                                                    & 2.31                                                      \\
$x^{*3}$ (CF)  & Steel    & 565.6                                                & ... & 3.22                                                            & 1.58                                                    & 2.71                                                      \\
% $x^{*4}$ (CF)  & Steel    & 601.7                                                & ... & 2.12                                                            & 1.61                                                    & 1.87                                                      \\
% $x^{*5}$ (CF)  & Steel    & 565.6                                                & ... & 3.35                                                            & 1.56                                                    & 2.82             \\                                        
\bottomrule
\end{tabular}
}
\label{q3}
\end{table}

Through these examples, we have demonstrated that MCD excels at handling multi-objective quantitative performance queries and can identify performance-enhancing design modifications in these settings. 
In our next example, we consider a scenario in which more subjective text and image requirements are provided instead of quantifiable performance requirements. 

\section{Modifying Parametric CAD Designs to Match Text Prompts or Images} ~\label{CS2}

\begin{figure*}
    \centering
    \includegraphics[width=1\linewidth]{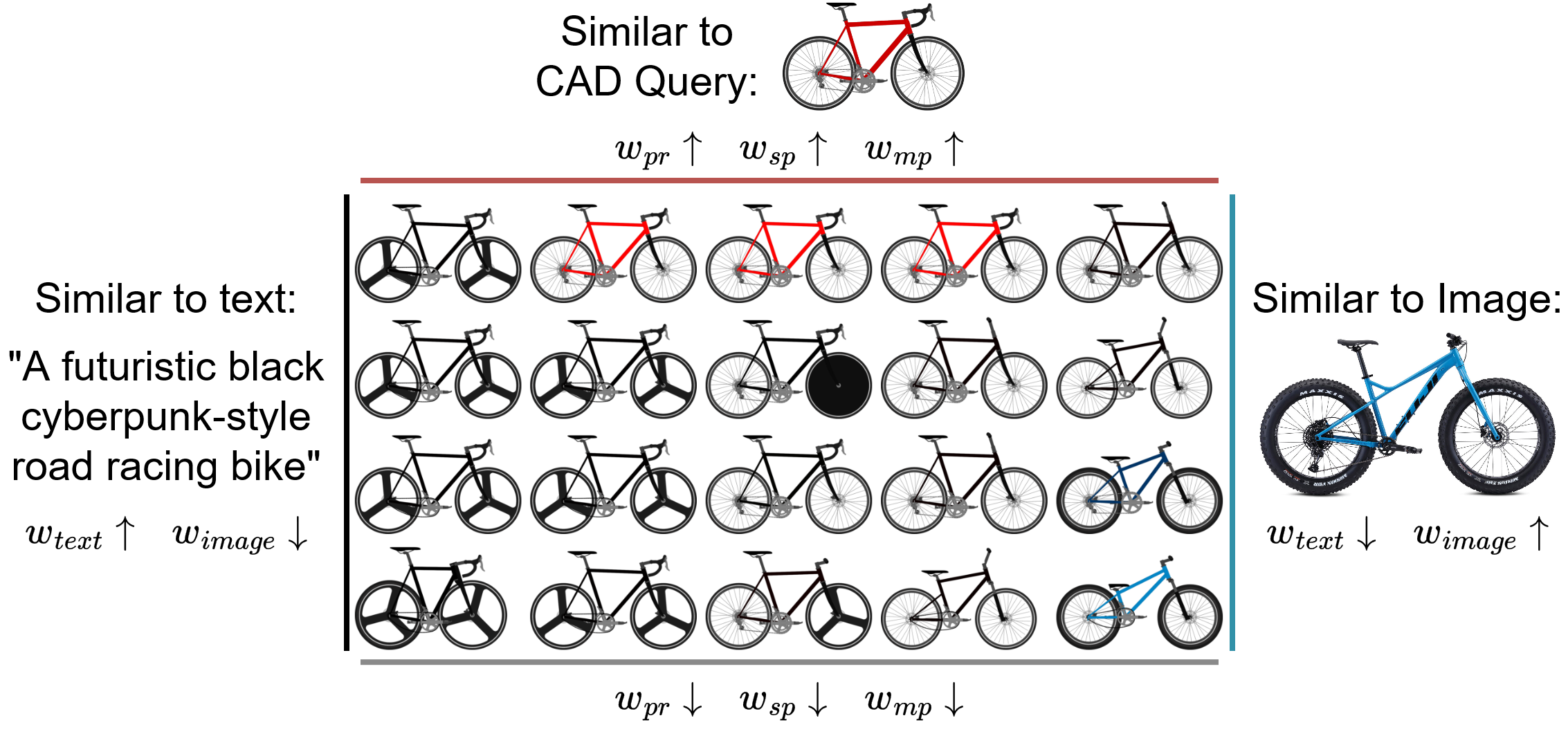}
    \caption{Visualization of the objective manifold for cross-modal counterfactual selection. Designs sampled from the top of the manifold prioritize proximity, sparsity, and manifold proximity. Designs on the left and right sides prioritize similarity to a text prompt and reference image, respectively. Note: Designs are \textbf{optimized parametrically} by modifying CAD features and are rendered for visualization purposes. Objectives are calculated in a cross-modal fashion.}
    \label{fig:case2}
\end{figure*}

In this case study we examine subjective counterfactuals like: ``What if my design looked more `cyberpunk'-themed?'' Classically, counterfactual search requires a clear and quantifiable counterfactual requirement. This can be constraining, since subjective counterfactual requirements may be more natural and intuitive for human users. This is often the case for images or text, which are much more easily understood by humans compared to tabular or parametric data. Accordingly, we demonstrate how we can query MCD using text prompts or reference images. Importantly, we still optimize designs parametrically, where design space variables directly correspond to features in a Computer-Aided Design (CAD) model. This can be seen as a cross-modal counterfactual search problem, where a CAD model is being modified to adhere to an image or text-based requirement. 

\subsection{Enabling Cross-modal Counterfactual Search}
To modify the user's current parametric CAD design to be more similar to a text or image, MCD needs an evaluator that can calculate the similarity between a parametrically-represented bicycle design and a text string or image. To do so, we use a model introduced in~\cite{regenwetter2024biked}, trained on 1.4 million bicycle designs to directly map parametric bike designs into a shared embedding space with text and images. Our optimization objectives are then calculated using a cosine similarity between the bike's embedding and the target text or image embedding. 

In this example, we select a subset of 97 variables from BIKED's parameter space as our design space, meaning that $x_k$ and $q_k$ for $k\in[1...97]$ are 97-dimensional parametric vectors. Objectives $f_{text}(x)$ and $f_{image}(x)$ are expressed as a function of a design, $x$, a reference text prompt, $T$, and reference image, $I$, as:
\begin{flalign}
    \begin{aligned}
    &f_{text}(x) = S_{cos}\left( E_p(x), E_t(T)\right), \\
    &f_{image}(x) = S_{cos}\left( E_p(x), E_i(I)\right),
    \end{aligned}
\end{flalign}
where, $E_p$ is the trained parametric embedding model from~\cite{regenwetter2024biked}, while $E_t$ and $E_i$ are the trained text and image embedding models from CLIP~\cite{radford2021learning}. $E_p(x)$, $E_t(x)$, and $E_i(x)$ are each 512-dimensional embedding vectors. The cosine similarity, $S_{cos}$ is expressed as:
\begin{equation}
    S_{cos}(A,B) = \frac{A\cdot B}{\mid A\mid\,\mid B\mid}
\end{equation}
We select a generic red road bike design from BIKED as our parametric query, $q$, the phrase ``A futuristic black cyberpunk-style road racing bicycle'' as our text prompt $T$ and an image of a blue Fuji Wendigo 1.1 mountain bike as our image prompt, $I$. In this context, the user is effectively asking questions like: ``How would I change my red road bike's parameters if I wanted it to look more like a black cyberpunk-style bike or like this blue mountain bike design?'' We optimize for 150 generations with a population size of 100. Next, we perform a series of sampling operations with different objective weights. By selecting the optimal bikes at a sweep of different objective weights, we can visualize the best bikes under numerous configurations of objective priorities, as shown in Fig~\ref{fig:case2}. Counterfactual quality objective weights in the $i^{th}$ row are chosen as:
\begin{equation}
    w_{pr} = w_{sp} = w_{mp} = 1.5^{2-i}
\end{equation}
In this way, counterfactuals with better proximity, sparsity, and manifold proximity are prioritized toward the top of the grid, while counterfactuals are given more leeway to deviate from the query design and data manifold toward the bottom. Diversity weight, $w_d$, was irrelevant, as only one design was sampled for each combination of objective weights. Similarly, auxiliary objective weights in the $j^{th}$ column were set in terms of the number of columns, $5$, as:
\begin{equation}
    w_{text}=1.5^{5-j},\,w_{image}=1.5^{j-1}
\end{equation}
These objectives allowed similarity to the first text prompt to take precedence on the left edge of the grid and similarity to the image prompt to take precedence on the right. 

\subsection{Cross-modal Counterfactual Results} 
As expected, when we render the CAD models at the top of the grid, they are appreciably similar to the red query design. Bikes further down the grid become progressively more visually different. As the counterfactual quality weights are relaxed toward the bottom, the counterfactuals are also more likely to have geometric incompatibilities or other types of ``implicit constraint'' violations. For example, the seat tube in the bottom left bike intersects slightly with the rear tire. 

Bikes in the lower left corner of the grid can be subjectively identified as more similar to ``A futuristic black cyberpunk-style road racing bicycle.'' Among the key modifications are a color change and a shift to tri-spoke wheels, which are arguably more on-theme for a `cyberpunk-style' bike. 

Likewise, bikes towards the bottom right corner of the grid can be subjectively identified as more similar to the blue mountain bike image. Bikes in this corner have the slanted down tube which is characteristic of mountain bikes, feature mountain bike handlebars and thicker tires, and even match the color of the image. 

In this case study we demonstrated that MCD can leverage cross-modal evaluators to optimize a design in a parametric design space to match text or image prompts in a cross-modal fashion. Next, we move on to consider a challenging multi-modal problem with six multidisciplinary quantitative and qualitative objectives as our final case study. 

\section{Modifying Designs with many Multidisciplinary Design Requirements} ~\label{CS3}
In this case study we tackle a fully multidisciplinary counterfactual search problem. We combine the qualitative aesthetics-focused requirements of the second case study with the quantitative structural-mechanics-focused requirements of the first case study. In addition, we introduce two new requirements focused on user-customizability. These requirements quantify aerodynamic drag on the cyclist and overall ergonomic `fit.' These functions are calculated as a function of both the bicycle design and the rider's body dimensions. 

\begin{figure*}
    \centering
    \includegraphics[width=1\linewidth]{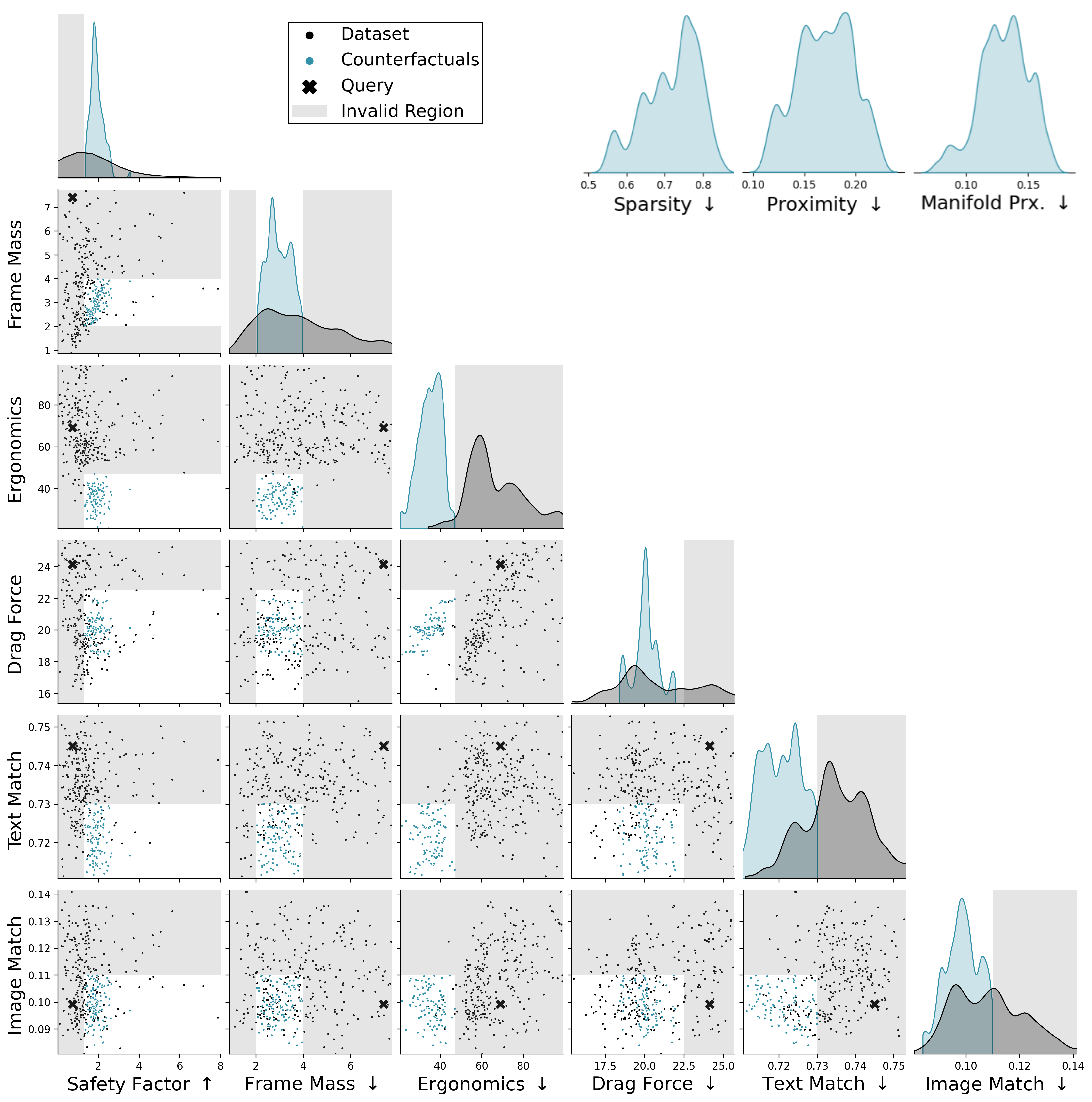}
    \caption{Pairplots visualizing objective score distributions for identified counterfactuals and the dataset. Individual kernel density estimates are shown on the diagonal, while pairwise scatterplots are shown on the off-diagonal. The counterfactual query's objective scores are marked with a black X. We also show sparsity, proximity, and manifold proximity score distributions over the counterfactuals.}
    \label{fig:pairplots}
\end{figure*}

\subsection{Configuring MCD with Many Requirements}
We summarize the counterfactual requirements as follows. The first four requirements and reference text and image prompts, $T$ and $I$, are reused from the previous application examples (Sec.~\ref{CS1} and~\ref{CS2}):
\begin{flalign}
    \begin{aligned}
    &f_1(x) = S_{cos}\left( E_p(x), E_t(T)\right) \\
    &f_2(x) = S_{cos}\left( E_p(x), E_i(I)\right) \\
    &f_3(x) = R_{mass}(x)\\
    &f_4(x) = R_{SF}(x)
    \end{aligned}
\end{flalign}
We add two new requirements $f_5(x)$ and $f_6(x)$ that are calculated as a function of a set of rider dimensions, $r$, comprised of five key body dimensions: lower leg length, upper leg length, torso length, arm length, and overall height. These functions are expressed as follows:
\begin{flalign}
    \begin{aligned}
    &f_5(x) = R_{drag}(x, r)\\
    &f_6(x) = G_{ergo}(x, r)\\
    \end{aligned}
\end{flalign}
Here, $R_{drag}$ is a model trained on the cyclist aerodynamic drag dataset introduced in~\cite{regenwetter2022data} to predict drag force on the cyclist in a direct $10\,m/s$ relative headwind. $G_{ergo}$ is a composite score calculating: 
\begin{equation}
    G_{ergo} = |\theta_{K} - \theta^*_{K}|\,+\,|\theta_{H} - \theta^*_{H}|\,+\,|\theta_{A} - \theta^*_{A}|
\end{equation}
Here, $\theta_{K}$, $\theta_{H}$, and $\theta_{A}$ are the rider's joint angles at the knee, hip, and armpit, respectively, while $\theta^*_{K}$, $\theta^*_{H}$, and $\theta^*_{A}$ are optimal reference values for the three angles sourced from~\cite{burt2022bike}. Joint angles $\theta_{K}$, $\theta_{H}$, and $\theta_{A}$ are derived analytically using a set of trigonometric relations included in the project's codebase. Rider body dimensions for this example were selected semi-randomly using a reference of anthropometric statistics. $L_i$ and $U_i$ are selected as indicated by the gray regions in Fig.~\ref{fig:pairplots}.

We select a starting query from the BIKED dataset and a challenging set of objective targets across the six objectives. These targets are so demanding that none of the $\sim4500$ existing designs in the dataset simultaneously achieve all six targets. 100 counterfactuals are generated as potential modifications to the query design.

% Like the last case study, we sample designs in a grid, as shown in Fig.~\ref{fig:multimodal} based on a variable objective weighting scheme. We select a spread of DTAI objective weighting parameter ($\alpha$) values in terms of the $i^{th}$ row and $j^{th}$ column as follows: 
% \begin{equation}
% \begin{split}
%     \alpha_{\text{text}}=2^{n-j},\,\alpha_{\text{image}}=2^{j}\\
%     \alpha_{\text{sf}}=1.5^{n-i-1},\,\alpha_{\text{mass}}=1.5^{i} 
% \end{split}
% \end{equation}
% This time, we hold the counterfactual quality objective weights constant at:
% \begin{equation}
%     w_{pr} = w_{sp} = w_{mp} = 0.05
% \end{equation}

\subsection{MCD's Performance with Many Requirements}
The sampled counterfactuals comfortably achieve all six design requirements. Figure~\ref{fig:pairplots} shows the objective score densities for each individual objective (diagonal) and pairwise scatterplots over every set of two objectives (off-diagonal). The generated counterfactuals, query, and dataset are plotted. The small white regions constitute the counterfactual requirements. We also plot the distribution of sparsity, proximity, and manifold proximity scores over sampled counterfactuals. Notably, MCD finds counterfactuals that change only half of the design parameters or change parameters by on average only 10\% compared to the span of the dataset. 

We stress that many of these objectives are challenging to simultaneously optimize. As discussed in the first case study, increasing the safety factor while reducing mass is a challenging structural optimization task. Simulataneously matching two very different subjective prompts is similarity challenging, as showcased in the second case study. Even the two additional objectives, ergonomics and aerodynamics, are challenging to simultaneously optimize since designing the bike to position riders in a highly aerodynamic stance may require extreme joint angles. As the case study demonstrates, MCD can effectively find solutions that balance this demanding set of requirements, creating many options for highly efficient bikes, specially customized for an individual user's biomechanical needs and a set of arbitrary subjective preferences. 

\section{Performance Analysis and Ablation}
    In this section, we contrast the performance and computational cost of counterfactual search using MCD to classic optimization. 
\subsection{Comparing MCD to Classic Optimization} \label{sec:ablation}
\begin{figure}[!htb]
    \captionsetup[subfigure]{justification=centering}
    \centering
    \begin{subfigure}[b]{\linewidth}
        \centering
        \includegraphics[width=\textwidth]{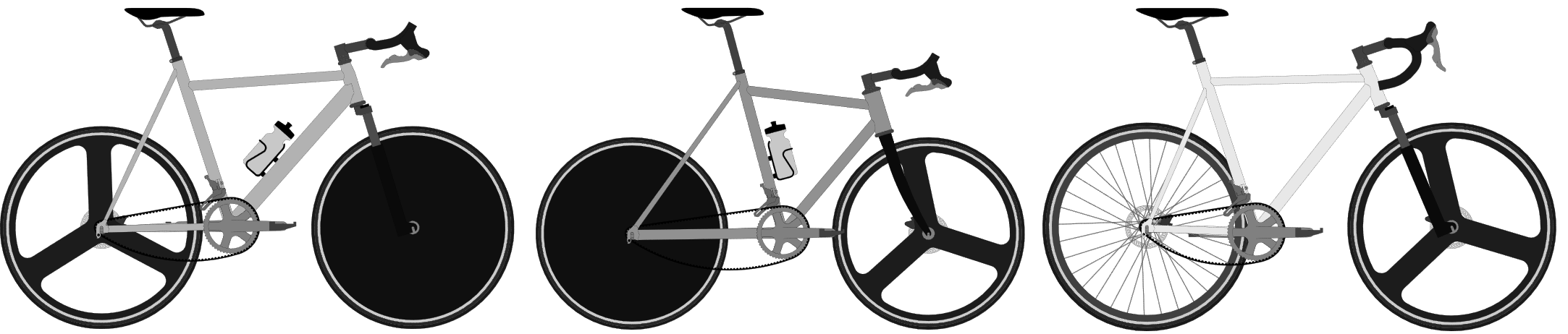}
        \caption{Counterfactual Search (MCD)}
        \label{fig:query}
    \end{subfigure}
    \hfill
    \begin{subfigure}[b]{\linewidth}
        \centering
        \includegraphics[width=\textwidth]{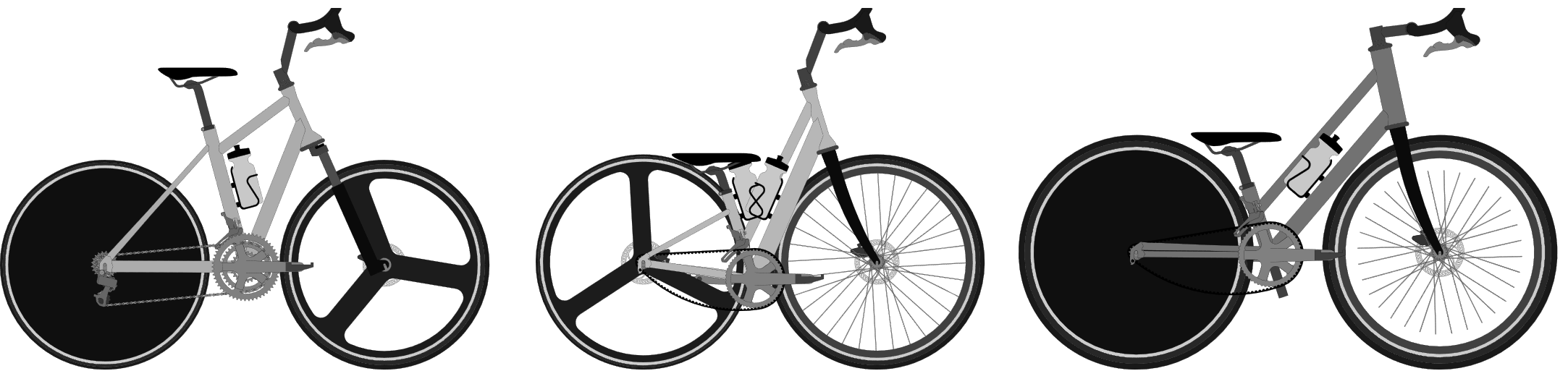}
        \caption{Optimization (NSGA-II)}
        \label{fig:dir}
    \end{subfigure}
    \caption{Comparison of bikes generated by MCD and bikes generated by classic optimization. Optimization is heavily reliant on explicit and exhaustive constraints and tends to fail in their absence. In contrast, counterfactual search can implicitly obey unknown constraints through its proximity, sparsity, and manifold proximity objectives. }
    \label{fig:ablation}
\end{figure}

Counterfactual search has several advantages over classic optimization. These advantages are largely enabled by the counterfactual quality attributes optimized in MCD --- sparsity, proximity, and manifold proximity. One key advantage is the ability to implicitly fulfill constraints in the absence of exhaustive and explicit constraint definitions. Whereas optimization tends to generate invalid designs that violate implicit constraints, counterfactual search has a much higher chance of satisfying constraints. This is because it places a priority on making minimal changes to the query design, which increases the likelihood of generating valid counterfactuals. Furthermore, manifold proximity encourages MCD to select designs that lie within a data manifold of valid designs. We demonstrate this advantage over classic optimization in Figure~\ref{fig:ablation}. For this figure, we give the same optimization problem, adapted from the structural optimization case study, to both MCD and a plain optimization (NSGA-II) algorithm. The designs generated by NSGA-II are littered with intersecting geometry and functional issues (such as pedals impeding the steering of the front wheels), which are apparent when the designs are rendered. Though MCD's designs have a few problems, they are much more viable overall, and one design does not have any obvious issues. 

To quantitatively substantiate this argument, we evaluate MCD against NSGA-II but withhold the constraint checks provided in BIKED. We only use the constraints to later evaluate the validity of generated designs. As shown in Table~\ref{tab:ablation}, designs identified by MCD violate on average 0.92 constraints versus 1.51 for designs generated by NSGA-II. Collectively, MCD can identify valid designs that do not violate any constraints 35\% of the time to NSGA-II's 15\%. These scores are averaged over three tests, with standard deviations shown in the table.
\begin{table}[]
\centering
\caption{Counterfactual search significantly outperforms classic optimization in constraint satisfaction. MCD can generate more than double as many valid designs as NSGA-II. Constraint violation and valid design fraction scores are shown in the table, averaged over three runs, with standard deviations shown.}
\label{tab:ablation}
\resizebox{\linewidth}{!}{%
\begin{tabular}{lcc}
\toprule
& \begin{tabular}[c]{@{}c@{}}CF Search\\ (MCD)\end{tabular} & \begin{tabular}[c]{@{}c@{}}Optimization\\ (NSGA-II)\end{tabular} \\
\midrule
Constraint Violation (Avg. \#) $\downarrow$ & \textbf{0.92±0.03}                  & 1.51±0.12         \\
Fraction of Valid Designs $\uparrow$      & \textbf{0.35±0.04}                  & 0.15±0.05        \\
\bottomrule
\end{tabular}%
}
\end{table}

\subsection{Timing MCD's Computational Requirements}
MCD's additional functionality adds modest computational cost as compared to conventional optimization. We calculate the cumulative computational cost for the counterfactual search performed in the final (multidisciplinary) case study and sort major components by type. Any cost directly incurred through the evaluation of the six objective scores for a design is considered part of the ``evaluator'' cost. Any computation that is introduced by MCD (would not be performed during a classic optimization run) is associated with MCD. Any other cost is attributed to the optimizer itself. Both the optimizer and MCD's costs (9 and 20 seconds, respectively) are dwarfed by the evaluation functions (375 seconds), despite all evaluation functions being fast predictive models or analytical equations in this study. Since 15K design evaluations are performed, MCD adds just over 1 ms per design evaluation. We note that computational cost is dependent on a variety of factors like population size and dataset size. Regardless, converting an optimization process into a counterfactual search using MCD should generally be a lightweight modification. 
\begin{figure}
    \centering
    \includegraphics[width=1\linewidth]{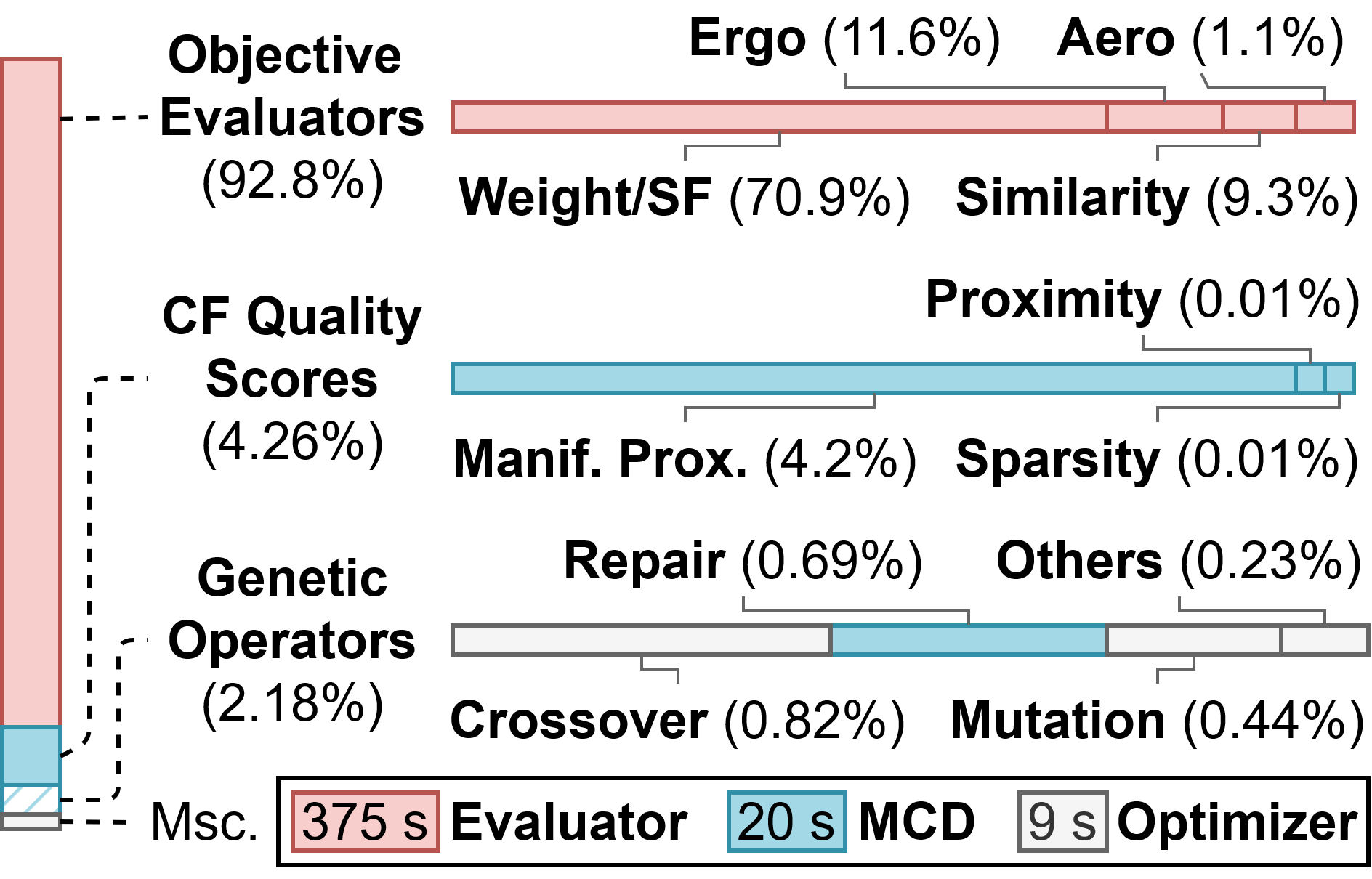}
    \caption{Computation time breakdown for the final multidisciplinary counterfactual search example. Functions are sorted and color-coded as evaluation functions, features of the underlying optimization, or specific features of MCD. MCD's functions take on the same order of time as the internal optimizer's functions. Even though the evaluators are all fast predictive models or analytical equations, they still take the overwhelming majority of the computational expense. }
    \label{fig:enter-label}
\end{figure}
\section{Limitations}
MCD makes several key contributions to counterfactual search, primarily targeting designers. However, it also has a few limitations. For example, MCD's model agnosticism allows it to support nondifferentiable evaluators but comes at the cost of potentially lower sample efficiency compared to gradient-based methods. Another key limitation stems from the difficulty of genetic algorithms in handling a large number of objectives. Because MCD adds three counterfactual quality objectives to the objective space, it slightly exacerbates the dimensionality issue of multi-objective genetic algorithms. Future work will explore MCD variants that leverage gradient information and many-objective optimization methods to address these limitations.   

Finally, we note that the CLIP-based text- and image-matching approach is effective at capturing abstract and subjective aesthetics, but may struggle to capture technical details supplied in reference prompts. As such, we recommend that users with highly technical constraints specify them parametrically, instead of through text. However, as machine learning models continue to improve, querying counterfactual models for precise technical details through text and images may improve significantly. 

% \section{Discussion}

\section{Broader Impact}
Design tools like MCD have the power to both accelerate and democratize design. By recommending high-performing design modifications, MCD can save designers many rounds of painstaking trial and error. This would enable designers to rapidly iterate and accelerate their design workflows. Cross-modal counterfactual search can also enable designers to interact more intuitively with computational design tools by specifying design requirements through text and images. This may lower the bar of entry for prospective designers who would lack sufficient expertise to rigorously define and constrain optimization problems. In short, we are optimistic that MCD takes a step towards more powerful and intuitive computational design tools. 

\section{Conclusion}
In this paper, we have introduced Multi-objective Counterfactuals for Design (MCD), a specialized counterfactual search method for design modification tasks. We first illustrated how design modification can be addressed through counterfactual search and what attributes comprise a strong design modification. We then identified key limitations with existing works, particularly their inability to sample multi-objective queries and the inherent coupling of the optimization and sampling process. Next, we demonstrated how MCD solves these two challenges using several case studies from the challenging real-world engineering problem of bicycle design. We first showed how MCD's support for multi-objective queries allowed it to recommend meaningful modifications to improve the structural efficiency of bicycle frames. We then showed that MCD can leverage cross-modal evaluators to identify counterfactuals that satisfy highly subjective requirements specified through images or text prompts. We finally showed that MCD can individually customize design modifications (such as for a user's biomechanical attributes), while simultaneously optimizing for quantitative performance requirements and subjective design requirements. 

All in all, MCD is a intuitive computatonal design tool that can quickly recommend design modifications customized to fulfill both quantitative and subjective design requirements. We are excited to release our code, demos, and experiments at \url{http://decode.mit.edu/projects/counterfactuals/} and anticipate a variety of interesting use cases across the community. 
\bibliographystyle{IEEEtran}
\bibliography{bibliography}
\appendix
\section{Description of BIKED and Auxiliary Datasets}
This paper builds several case studies on the BIKED dataset~\cite{regenwetter2022biked} and associated datasets which interface with and augment the main dataset. The BIKED dataset is composed of ~4,500 community-designed bicycle designs. It is a multi-modal dataset: each design is represented parametrically, as a set of component images, and as an assembly image. The parametric representation is quite detailed, with more than 23,000 dimensions in the 'raw' dataset. In the case studies presented, we use smaller subsets of the full parametric design space. These parametric representations directly correspond to features from BikeCAD models and can be quickly rendered using a tool introduced in~\cite{regenwetter2024biked}. 
In the first study, we leverage the FRAMED dataset~\cite{regenwetter2023framed}, which expands on BIKED with structural analysis data from finite element simulations of ~4,000 human-designed bikes and ~10,000 AI-generated bikes. In the second case study, we additionally use BIKED++~\cite{regenwetter2024biked}, a dataset of 1.4 million procedurally-generated bicycle designs represented parametrically and as images. Finally, in the third case study, we use an aerodynamic drag predictor from~\cite{regenwetter2022data} trained on finite element simulations of 4000 cyclist configurations. 
\end{document}